\pdfoutput=1
\documentclass[9pt,journal,compsoc]{IEEEtran} 
\ifCLASSOPTIONcompsoc
  \usepackage[nocompress]{cite}
\else
  \usepackage{cite}
\fi
\ifCLASSINFOpdf
   \usepackage[pdftex]{graphicx}
\else
\fi

\usepackage{amsmath}
\usepackage{placeins}
\usepackage{multirow}
\usepackage{balance}
\usepackage{amsfonts}
\usepackage[ruled,vlined]{algorithm2e}
\usepackage{float}

\usepackage{ragged2e} 
\begin{document}

%
\title{NeuRegenerate: A Framework for\\
Visualizing Neurodegeneration}
%
%
%
%

\author{Saeed Boorboor, 
Shawn Mathew, 
Mala Ananth, 
David Talmage, 
Lorna W. Role, 
and Arie E. Kaufman,~\IEEEmembership{Fellow,~IEEE}
\IEEEcompsocitemizethanks{\IEEEcompsocthanksitem Boorboor, Mathew, and Kaufman are with the Department of \protect\\
Computer Science, Stony Brook University.\protect\\ 
E-mail:~\textit{\{sboorboor, shawmathew, ari\}}@cs.stonybrook.edu

\IEEEcompsocthanksitem Ananth, Talmage, and Role are with the National Institutes of Health.}
\thanks{Manuscript received xx xxx. 202x; accepted xx xxx. 202x. Date of Publication xx xxx. 202x;}}

\IEEEtitleabstractindextext{%
\begin{abstract}
\justifying
Recent advances in high-resolution microscopy have allowed scientists to better understand the underlying brain connectivity. 
However, due to the limitation that biological specimens can only be imaged at a single timepoint, studying changes to neural projections over time is limited to observations gathered using population analysis. 
In this paper, we introduce {\em NeuRegenerate}, a novel end-to-end framework for the prediction and visualization of changes in neural fiber morphology within a subject across specified age-timepoints.
To predict projections, we present {\em neuReGANerator}, a deep-learning network based on cycle-consistent generative adversarial network (GAN) that translates features of neuronal structures across age-timepoints for large brain microscopy volumes. 
We improve the reconstruction quality of the predicted neuronal structures by implementing a density multiplier and a new loss function, called the hallucination loss.
Moreover, to alleviate artifacts that occur due to tiling of large input volumes, we introduce a spatial-consistency module in the training pipeline of neuReGANerator. 
Finally, to visualize the change in projections, predicted using neuReGANerator, NeuRegenerate offers two modes: (i) {\em neuroCompare} to simultaneously visualize the difference in the structures of the neuronal projections, from two age domains (using structural view and bounded view),
and (ii) {\em neuroMorph}, a vesselness-based morphing technique to interactively visualize the transformation of the structures from one age-timepoint to the other. Our framework is designed specifically for volumes acquired using wide-field microscopy. We demonstrate our framework by visualizing the structural changes within the cholinergic system of the mouse brain between a young and old specimen. 
\end{abstract}

}

\maketitle

\IEEEdisplaynontitleabstractindextext

%
\IEEEpeerreviewmaketitle

\IEEEraisesectionheading{\section{Introduction}\label{sec:introduction}}

\IEEEPARstart{T}{he} field of connectomics~\cite{TheHumanConnectome}, has been one of the major scientific endeavours of the 21\textsuperscript{st} century, with its goal to reconstruct a complete structural and functional connectivity of the brain across multiple scales of spatial and temporal resolution. 
Advancements in microscopy, particularly the potential to acquire high-resolution 3D images of the brain coupled with novel visualization techniques~\cite{pfister2014visualization,Awami14,Mohammed18,boorboor2018visualization,wan2012fluorender}, have provided neuroscientists with the tools to reconstruct and better understand brain connectivity. 
As a result, findings aided by these methodological advancements have challenged previous knowledge of system structure and organization.
One such system is the cholinergic system.
Since the early 1970s, alterations to the cholinergic system have been considered hallmarks of late-stage cognitive impairment. 
While focus initially has been on the loss of cell bodies, it has now become increasingly evident that neural projections -- the connections these neurons make with other regions of the brain -- are  more vulnerable and susceptible to fragmentation and loss, far before overt signs of impairment. 
To this end, neuroscientists are becoming increasingly interested in using visualization techniques to understand the morphological changes that occur to cholinergic fibers across lifespan. 

\begin{figure}
    \centering
    \includegraphics[width = \linewidth]{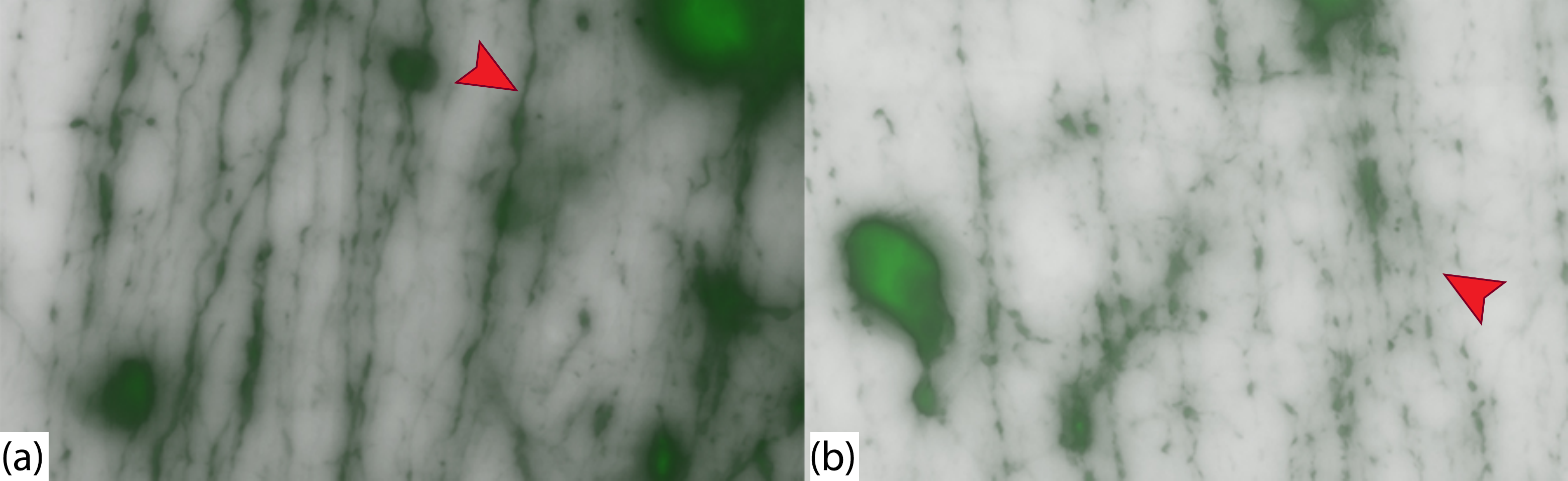}
    \caption{The axonal projections of (a) a 6-week, and (b) a 9-month mouse, imaged using WFM. We can observe that they appear connected and healthy at 6-weeks, and fragmented and thin at 9-months (red arrows). 
    }
    \label{fig:comparison_raw_original}
\end{figure}

The task of studying neuronal structures at micro-scales require physically sectioning brain samples, processing for cell-type specific labeling, imaging with a microscope, and employing visualization techniques for qualitative and quantitative evaluation.
Such a workflow means that samples can only be imaged at a single timepoint, thereby forcing researchers to use population analysis to study changes in neuronal structure, across lifespan.
That is, observations can only be made about general changes in the state of the fibers across timepoints across subjects~(Fig.~\ref{fig:comparison_raw_original}).
What is needed is a more precise understanding and a complete visualization of the profile of neurites at an individual level -- \textit{within-subject analysis}.

To this end, we present {\em NeuRegenerate}, a novel framework that predicts and reconstructs neuronal structures of an individual specimen, across specified age timepoints-of-interest.
NeuRegenerate has two main modules: a deep-learning module called {\em neuReGANerator} and a visualization module. 
Based on their usual protocol, neuroscientists image a brain region from a population of animal subjects, at the ages-of-interest under study.
Using this collection of wide-field~(WF) microscopy volumes, neuReGANerator is designed to learn the structural features of neurites that translate across the age domains.
Since the nature of this workflow lacks paired ground-truth correspondence between age domains, we utilize extended cycle-consistent generative adversarial networks (XDCycleGAN)~\cite{mathew2020augmenting} as the underlying model.
For our task at hand, to achieve a volume-to-volume translation specifically for brain microscopy data, we have designed neuReGANerator to consist of 3D convolutional neural networks, a density multiplier to emphasize the reconstruction of neuronal structures over background voxels, and a hallucination loss to avoid phantom data generation.
Moreover, we introduce a novel spatial-consistency architecture in our training pipeline to facilitate the learning of connectivity information of fiber structures and alleviate tiling artifacts in the model output volume.
NeuReGANerator is evaluated using qualitative assessments by domain scientists.

The inherent blurring nature of WF microscopy~(WFM) data makes effective and meaningful visualization of neuronal data a challenging task.  
To visualize changes in neurite structure across age timepoints, reconstructed using a trained neuReGANerator, our framework provides two visualization modes: neuroCompare and neuroMorph.
NeuroCompare, is a visualization that simultaneously compares the fiber profiles of the input and predicted brain samples, using both structural and bounded representations. 
For the structural representation, we encapsulate neuronal structures extracted from the old-age domain inside translucent iso-surfaces of their corresponding young-age structures. 
To achieve this, we also address the challenge in extracting surfaces of neuronal data from WFM volumes. 
In the bounded representation, neuroscientists can observe the structural representation of predicted neurites with respect to direct volume rendering of the original input volume. 
For a more interactive visualization, neuroMorph, a vesselness (tubular-shaped morphology) based volume morphing technique, reconstructs meaningful transformations between the original volume and the neuReGANerator output.

NeuRegenerate is designed based on the goals provided by our neuroscientist team members. 
The results and visualizations presented in this paper use data that are currently being used to study the fragmentation and loss of cholinergic fibers between 6-week (young) and 9-month old (aged) mice, for the medial septum region of the brain.
The contributions of this paper are as follows:
\begin{itemize}
    \item To the best of our knowledge, the first novel framework for the prediction, reconstruction, and visualization of changes that occur in the neuronal profile of an individual specimen;
    \item A deep-learning model specific to learning neuronal features from brain optical microscopy volumes; and 
    \item A vesselness-based volume morphing technique for visualizing changes in neurite projections across lifespan.
\end{itemize}

\section{Related Works}
\textbf{Deep-learning for neuronal data. }
Deep learning has shown promising results for tasks such as neuronal structure segmentation \cite{wang2019segmenting,turaga2010convolutional,cohen2018distribution,li2017deep}, neuron tracing~\cite{gala2014active,chen2015smarttracing}, and data generation \cite{wu2019three}.
For using neural networks for classification,  Zhou et al.~\cite{zhou2018deepneuron} have developed DeepNeuron, a toolbox for neuron tracing and analysis.
For reconstructing densely labelled axons in 2-photon microscopy, Skibbe et al.~\cite{skibbe2018pat} have proposed a probabilistic axon tracking algorithm in local-to-global model. 
Li et al. \cite{li2018large} have described a network to extract features for easy navigation though neuron databases. 
To tackle large microscopy volumes processing, Liu et al. \cite{liu2017detecting} have presented a method for training a network on gigapixel images. 
For synthesizing 3D volumes, Wu et al.~\cite{wu2019three} have proposed Deep Z, a digital image refocusing framework for fluorescence microscopy using a trained deep neural network to digitally reconstruct 3D samples using a single 2D WF image.

One important, common challenge faced by these proposed works is the large size of microscopy data that limits GPU-accelerated deep-learning tasks. Common solutions include cropping, tiling, or transforming the volume into 2D space. In NeuRegenerate, we present a novel approach in our deep-learning model for processing cropped tiles that alleviates artifacts due to tiling and significantly improves the quality of reconstructions by considering local neighborhoods. 
Existing works focus mainly on extracting neuronal data from microscopy volumes. To the best of our knowledge, no work uses deep-learning to reconstruct and predict neuronal structures across different timepoints of the lifespan from microscopy volumes at micrometer resolution.

Recently, deep-learning  has been used for image-to-image translation, such  as style transfer and image enhancement. 
Pix2pix~\cite{isola2017image}, for example, is a deep learning model that combines an L1 loss with a conditional GAN~\cite{mirza2014conditional} to handle image-to-image domain translation. This model requires paired groundtruth data from two given domains, which is not available in our context.
To address this constraint, unpaired image-to-image domain translation approaches including CycleGAN \cite{zhu2017unpaired} and like approaches \cite{kim2017learning,yi2017dualgan} are used.

\vspace{2pt}
\noindent
\textbf{Visualization methods. }
Techniques developed for the reconstruction, visualization, and analysis of complex neural connection maps have paved the way for neurobiologists to gain insights into the underlying brain structure and function. 
Volume renderers have been developed for 3D reconstruction and visualization of brain microscopy images. Mosaliganti et al.~\cite{mosaliganti2008reconstruction} have developed methods for axial artifacts correction and 3D reconstruction of cellular structures from optical microscopy. Nakao et al.~\cite{nakao2014visualizing} have presented an interactive visualization and proposed a transfer function design for 2-photon microscopy volumes based on feature spaces. Wan et al.~\cite{wan2012fluorender} have developed interactive rendering for confocal microscopy data that combines multi-channel volume rendering and polygon mesh data.  
For an immersive approach, Usher et al.~\cite{usher18} have introduced virtual-reality for the tracing of neurons.
Due to blurred nature of WFM data, simply applying volume rendering does not yield effective neurons visualization. Janoos et al.~\cite{janoos2008classification} have presented surface representation for the reconstruction of neuronal dendrites and spines from optical microscopy. Their solution, however, is meant for 2-photon microscopy, which does not exhibit blurring challenges as in WFM.

While we directly visualize the change in neuronal structures across age, a large body of
work is in time-dependent volume visualization.
Widanagamaachchi et al.~\cite{widanagamaachchi2012interactive} have employed feature tracking graphs.
Fang et al.~\cite{fang2007visualization} have analyzed volume differences for medical applications.
Lu and Shen~\cite{lu2008interactive} have proposed interactive storyboards composed of volume renderings and descriptive geometric primitives. 
Limited to a narrow range of time series to visualize our predictions, we derive motivation from Frey and Ertle~\cite{frey2016progressive} who have introduced a volume morphing technique and demonstrate its utility in reconstructing a full sequence for temporal data.
A large body of work is available for automatic volume morphing techniques~\cite{he1994wavelet,bai2014registration,fang2000volume,correa2010constrained}. While these works are designed to produce a morphing schedule for a source to a target volume, our proposed algorithm in neuroMorph morphs an entire set of neurites in the volume. Moreover, neuroMorph takes into account the specific vesselness property of the neurites to assign morphing paths for in-between volumes.

\section{Domain Background and Goals}
\label{sec:goals}
Along with other neuroscientists, we have recently observed that cholinergic fibers are vulnerable early-on in aging, and susceptible to loss far sooner than previously understood.
These fibers originate from the cell body and project widely across the brain, propagating information from one region to another and facilitating coordination to support both simple and complex behaviors.
Given this, a large focus of our work has been devoted to identify vulnerable and resilient regions of the cholinergic system and the progression of changes that occur to fiber integrity during aging.
An attempt to do this is by using diffusion tensor imaging or diffusion spectrum imaging. 
These techniques, though useful for repeat investigation in the same subject, provide broad information about tracts or bundles of fibers, whereas the changes we expect early on in aging are at a finer scale of individual fibers. Visualization of these fibers across brain regions and across lifespan will help neuroscientists understand the relationship between changes to the cholinergic system and changes to cognition.
Thus, we define the following goals for NeuRegenerate development:   
\paragraph*{\textbf{G1}}  
A framework that can predict a plausible morphology of cholinergic fibers of an individual specimen brain sample, for a future or past age timepoint. While we can examine fibers at discrete timepoints, it is challenging to get a sense of the individual differences in fiber density within subjects, given the limitations presented by standard workflows. Specific to neurodegeneration, this accounts for reconstructing fiber thickness and continuity due to fragmentation. Moreover, the framework should account for differential vulnerability or resilience across regions of the brain, as observations suggest fibers are affected at different rates in different regions.
\paragraph*{\textbf{G2}}
A visualization application that can provide neuroscientists with an intuitive method for analyzing the extent of changes in fiber morphology between the input microscopy volume and the reconstructed output predicted by the framework defined in \textbf{G1}. 
Particularly, based on the motivation behind the scientific investigation, visualizing fiber morphology becomes more significant than rendering voxel intensity values of raw microscopy volumes. 
Moreover, a time-series representation depicting the progression of the structural changes across timepoints will be useful to visualize the data in a continuous manner, not only for experts, but also to intuitively demonstrate predicted results to non-experts.

\paragraph*{\textbf{G3}}
The framework should support WFM data as input. WF~\cite{WFMintro} is a type of fluorescence microscope, widely used as primary imaging modality for experimental investigations. Its popularity is mainly due to large imaging field-of-view and faster imaging time~(compared to light-sheet and confocal microscopes). Because of these two features, imaging consecutive slices, rather than subsampling (as commonly done in confocal imaging) is standard, allowing for acquisition of continuous information.
However, WF data suffer from degraded contrast between foreground and background voxels because of out-of-focus light swamping the in-focus information, low signal-to-noise ratio, and poor axial resolution.
To this end, the prediction component of the framework should be resilient towards background noise and the visualization components should address the challenges of rendering WFM data.

\section{NeuRegenerate Framework}

\begin{figure}[t]
    \centering
    \includegraphics[width = \linewidth]{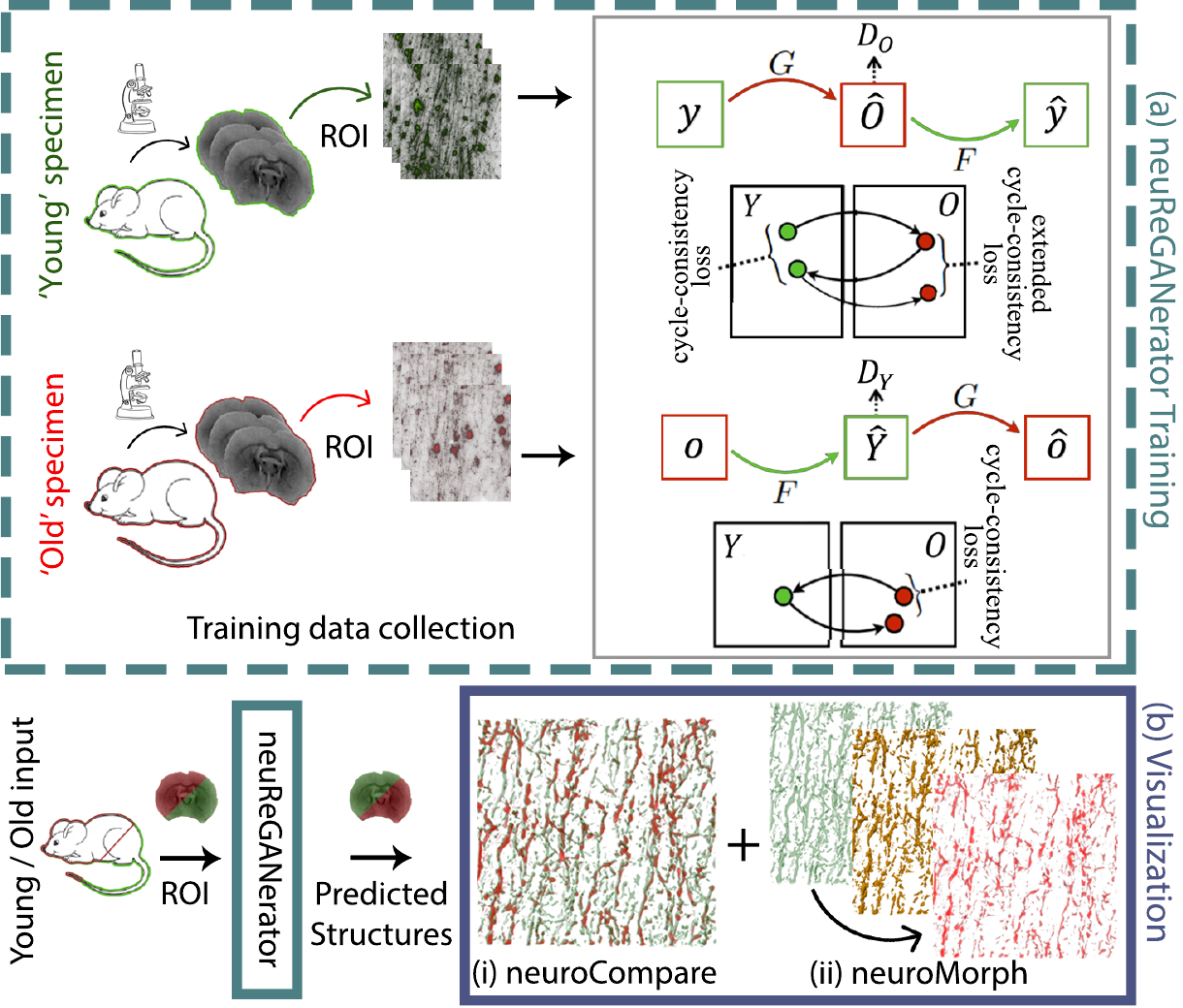}
    \caption{ NeuRegenerate framework for (a) training and (b) visualization.
    By providing a collection of region-of-interest WFM brain volumes, neuReGANerator learns the structural features that are translated between the specified age timepoints.
    Once trained, neuReGANerator predicts fibers for the respective age translation. We provide 2 visualization modes: (i) neuroCompare: simultaneous visualization of predicted result and input data (using structural view and bounded view), and (ii) neuroMorph: interactive transformation of input volume to predicted result using our morphing technique. }
    \label{fig:pipeline}
\end{figure}

Based on the requirements defined by our neuroscientist collaborators, NeuRegenerate consists of two parts: (1) deep-learning model, neuReGANerator~(Sec.~\ref{sec:neureganerator}); and (2) using the trained model to predict, reconstruct, and visualize the neuronal profile changes between the age timepoints-of-interest~(Sec.~\ref{sec:vislualization}).
Fig.~\ref{fig:pipeline} illustrates an overview of this framework.

\subsection{Deep-Learning Network neuReGANerator}
\label{sec:neureganerator}
The current inability to sample fiber morphology at a micrometer resolution for multiple timepoints, as subjects have to be sacrificed and sectioned for imaging, makes the use of deep-learning methods for predicting fiber structural changes a viable and necessary solution.
In recent times, due to their ability to solve image-to-image translation tasks, 
GANs~\cite{goodfellow2014generative} are becoming a popular deep-learning tool for the synthesis and transformation of medical data across different domains and imaging modalities~\cite{nie2017medical,costa2017end,emami2018generating}.
GANs are networks that learn patterns in the data by training two models simultaneously: a \textit{generator}  that captures the characteristics of a training dataset and a \textit{discriminator}  that distinguishes between
 samples from the training data and those generated by the \textit{generator}.
The two models are trained in a zero-sum game until the \textit{generator} learns to reconstruct data that the \textit{discriminator} cannot distinguish from the original dataset.
In principle, GANs reconstruct samples by learning the distribution of the output domain, however, for tasks where there is a meaningful correspondence between two domains, GANs are structured in a conditional setting~\cite{isola2017image}.
A conditional GAN uses paired groundtruth correspondence between two specified domains to learn a unique solution that translates features in the input domain to a sample with features of the output domain.

Paired ground-truth correspondence can be difficult or expensive to obtain, or in our case, impossible.
To overcome this, cycleGAN~\cite{zhu2017unpaired} is designed to learn the bidirectional translation of image domains in the absence of paired training examples. CycleGAN captures the distinct characteristics of the image domains and learns to  translate those characteristics across the domains.
It consist of two GANs, $F: X \to Y$ and $G: Y \to X$, that are trained simultaneously using a cycle consistency loss.
Cycle consistency enforces that an image should be reproducible when translated to a different domain and reverted back.
That is to say, in mapping an element from one domain to the other, and back, the model should reconstruct the original element: $x \to G(x) \to F(G(x)) \approx x$ and $y \to F(y) \to G(F(y)) \approx y$.
Additionally, an adversarial loss ensures that the translated images reside in the corresponding data distribution. 
In some cases, an input domain is lossy (a domain which has multiple solutions) and thus XDCycleGAN~\cite{mathew2020augmenting} enforces the network to better understand the lossy domains, by introducing an extended cycle consistency loss in the lossy domain.

This motivates neuReGANerator, a volume-to-volume translation model, using XDCycleGAN, that learns the relationship of neuronal features across specified age timepoints and, as a result, visualize changes within individual specimen.
Fig.~\ref{fig:pipeline}(a) illustrates this concept.
As specified in \textbf{G1}, for any neurobiological system, the degradation of fiber health and thickness varies in different regions of the brain.
Therefore, to train neuReGANerator, a collection of a particular region of the brain, from specimens of two age timepoints-of-interest have to be prepared.
We use the terms \textit{young}~($Y$) and \textit{old}~($O$) in this paper to refer to the relative ages.

NeuReGANerator is designed specifically for optical microscopy brain volumes.
Our work addresses known issues with cycleGANs that hallucinate and remove features ~\cite{cohen2018distribution} and embeds information to trick the loss functions~\cite{chu2017cyclegan}.
To this end, we describe a density multiplier (Sec.~\ref{sec:density_multiplier}) and a new loss function, the hallucination loss (Sec.~\ref{sec:hallucination_loss}).
Moreover, due to limited memory bandwidth of commodity GPUs, input sizes for GPU-accelerated deep-learning models become restricted, especially for processing volumetric data. 
To alleviate artifacts that occur from tiling large input volumes, we introduce a novel spatial-consistency architecture (Sec.~\ref{sec:spatial_consistency}). 
Lastly, in Sec.~\ref{sec:GAN_implementation} we mention modifications that aid in training the network and significantly improve the quality of the reconstructed results.

\subsubsection{Density Multiplier}
\label{sec:density_multiplier}
The loss function enforcing cycle-consistency constraint, $\mathcal{L}_{cyc}$, is an $L1$ loss, that uses the mean difference between the original and reconstructed images.
Depending on the region of the brain, the ratio of foreground to background voxels vary.
While training, if samples have more background voxels than foreground voxels, then in an effort to quickly minimize the loss function, the model concentrates to learn the reconstruction of the background.
This is because background intensities are easy to learn and due to a larger voxel count their contribution to the loss function is significant.
Along such a learning trajectory, if the model encounters a data sample with more foreground than background voxels, the loss value skyrockets.
Such occurrences lead to heavy fluctuations in the cycle-consistency loss and as a result the network struggles to converge.

One possible solution is to tune the network's parameters and weights to help it converge.
However, our goal is to design a model that requires minimal parameter-tuning. 
Therefore, we introduce a dynamic weighting method for the cycle-consistency loss based on the density of the volume.  
It is common in classification tasks to put higher weights on samples of underrepresented classes. Similarly, we apply this concept on a voxel basis. We classify voxels based on a threshold and use the number of foreground voxels to determine the weight for individual samples. In other words, the density of a sample delegates its contribution to the loss.

For a domain $A$, the cycle-consistency loss is calculated using:
\begin{equation}
\begin{split}
     \mathcal{L}_{cyc}(G,F,A) &= \mathbb{E}_{a\sim p_{data}(A)} ~ \delta\Big(a,F(G(a))\Big) \cdot \| F(G(a)) - a\| \\
     \end{split}
\end{equation}
where  the density multiplier, $\delta$, is defined as: 
\begin{equation}
\delta(a,a_{rec}) = \frac{1}{totalVoxels - sharedBackground (a,a_{rec})}
\end{equation}

\subsubsection{Hallucination Loss}
\label{sec:hallucination_loss}
 
 To the best of our knowledge, there are no general approaches to resolve phantom artifacts generated by cycleGANs. 
 To address this issue, we introduce a hallucination loss that is novel for the task at hand. 
 Based on domain knowledge of neurodegeneration, we design this loss to penalize the network from generating phantom structures and restrict reconstructions to underlying neuronal structures in the input data. 
In neurodegeneration, all neuronal structures found in the old domain ought to exist in the young domain reconstruction. 
In reverse, neuronal structures reconstructed in the old domain should originate from structures present in the young domain. 
Assuming that $V_O$ is the old-domain volume, and $V_Y$ is the young domain volume, we introduce $\mathcal{L}_{hallucination}$ as:

\begingroup
\begin{equation}
    \mathcal{L}_{hallucination}(O,Y) = \mathcal{L}_{h_o}(O) + \mathcal{L}_{h_y}(Y)
\end{equation}

\begin{equation}
    \mathcal{L}_{h_y}(Y) =  \lambda_y \cdot \delta(V_{G(Y)},  V_Y) \cdot \sum min(max(V_{G(Y)} - V_Y,0),1) 
\end{equation}

\begin{equation}
    \mathcal{L}_{h_o}(O) = \lambda_o \cdot \delta(V_{G(O)},  V_O) \cdot \sum min(max(V_O - V_{F(O)},0),1)  
\end{equation}
\endgroup

\noindent
Essentially, this function masks intersecting voxels from both domains with a foreground signal. 
A loss is incurred proportional to all foreground voxels in the young domain that are not in the mask.    
By multiplying a desired weight, $\lambda$, to the hallucination loss, the network is penalized accordingly for violating the fact that all structures in the young domain should be composed of structures from the old domain.

\begin{figure}[t]
    \centering
    \includegraphics[width = \linewidth]{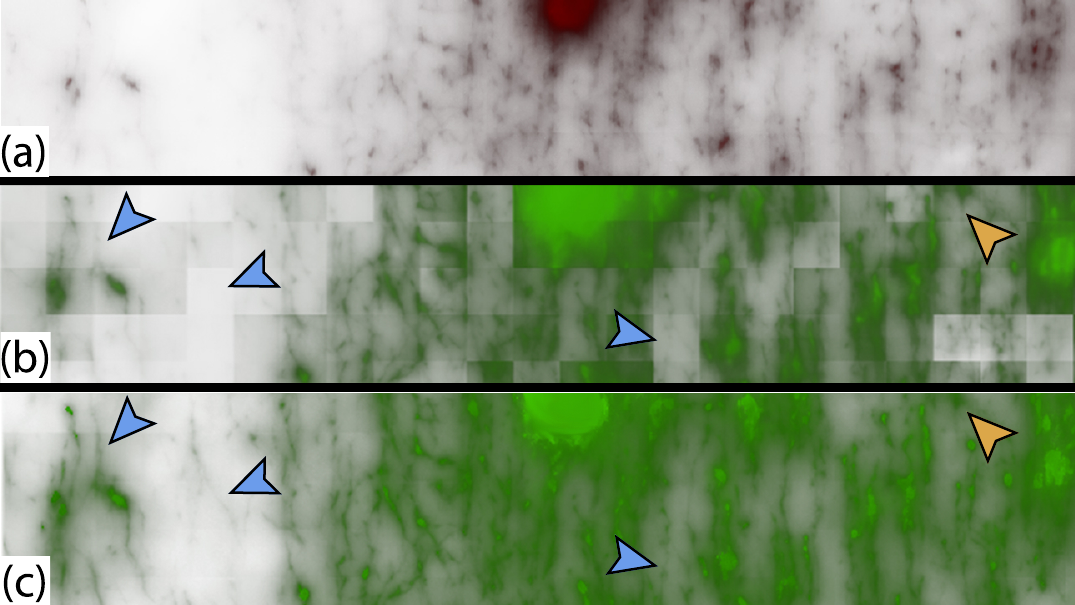}
    \caption{Our spatial-consistency module improves the reconstruction quality of the output volume. For an input raw WFM volume (a), the results without and with spatial-consistency are shown in (b) and (c), respectively. 
    We can observe that  in our results (c), tiling artifacts are reduced and the intensity values of foreground and background voxels across tiles are smoother and consistent, as pointed out by the yellow arrows. 
    Moreover, the reconstruction of neurites close to the tile borders are improved, as pointed out by the blue arrows.
    }
    \label{fig:comparison_spatial_result}
\end{figure}

\subsubsection{Spatial Consistency}
\label{sec:spatial_consistency}
\begin{figure}[t]
    \centering
    \includegraphics[width = \linewidth]{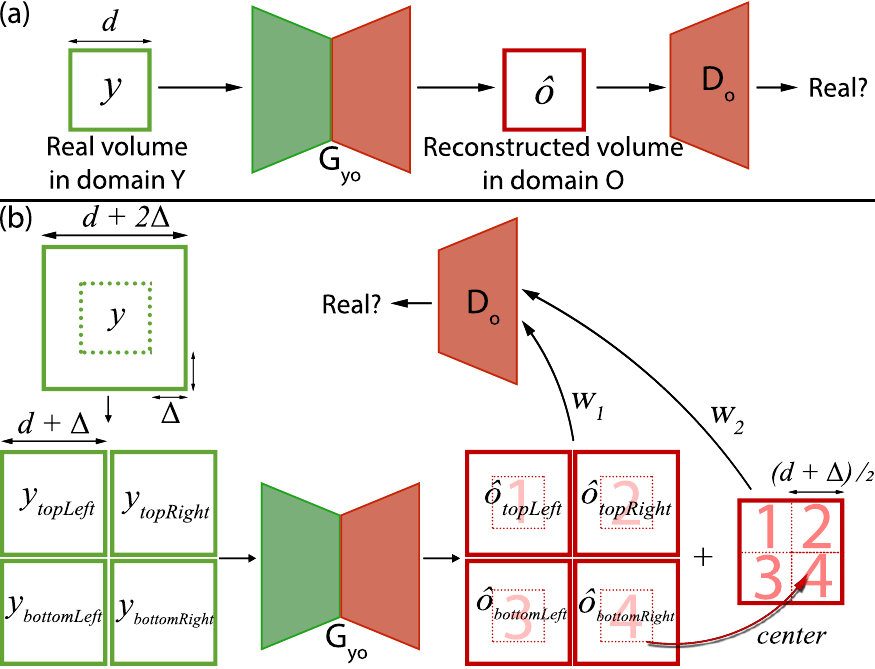}
    \caption{(a) Training a GAN. (b) Our spatial-consistency module, with simultaneous input of 4 overlapping tiles into the generator. The \textit{center} is constructed using the centers of each output from the generator.
    Using our spatial-consistency, the model is trained to consider information contained in a tile local neighborhood.}
    \label{fig:spatial_diagram}
    \vspace*{-2em}
\end{figure}
High resolution imaging of brain specimens result in volume data of extremely large sizes. 
For instance, a mouse brain slice with dimensions 7mm $\times$ 5mm $\times$ 20$\mu$m, imaged using WFM, yields a ~10 GB volume.
In practice, a large-sized input is either scaled to fit in GPU memory or the input data is tiled and processed individually. 
However, these approaches are not viable solutions for our task. 
Scaling down the input results in loss of data signals and cropping  isolates continuous information in the tile from its local neighborhood.

In the latter approach, a large input is divided into small tiles that can fit into the GPU memory for processing.
The choice of tile size becomes critical as they need to be large enough to contain the essential features that are necessary for the deep-learning network to properly training. 
Once trained, the input volume is divided into smaller, overlapping tiles.  
The output tiles from the network are then cropped from their center and stitched together to reconstruct the final output volume.
However, since the network processes each tile separately,  intensities of neuronal structures and background voxels vary across the output tiles, resulting in intensity inconsistencies and stitching artifacts in the final volume, as shown by the yellow arrows in Fig.~\ref{fig:comparison_spatial_result}(b).
Moreover, predicted neurites are reconstructed using input tiles that fully contain the features learnt by the network and are ignored otherwise, thus resulting in the partial reconstructions.

In order to alleviate the stitching artifacts and to produce better quality results, we introduce a spatial-consistency module in the training pipeline of neuReGANerator.
Using this approach, the network learns to include information in an input tile's local neighborhood .
Fig.~\ref{fig:spatial_diagram} illustrates the modifications we introduce in our network.
We explain the spatial-consistency module with a single direction, however, the concept is applied to both directions.

For an input volume with tiling dimensions $d \times d \times z$, we specify $\Delta$ as the neighborhood size. 
Therefore, a large-sized volume is divided into input tiles of size $(d + 2\Delta)^2 \times z$.
While training, neuReGANerator further divides the input tile into five smaller tiles, each of size $(d + \Delta)^2 \times z$: four overlapping quadrants - \textit{top left, top right, bottom left, bottom right}, and a \textit{center-}tile from the input tile centers.
The four quadrants are passed to the generator.
From the generator output, a tile of size $(\frac{d + \Delta}{2})^2 \times z$  is cropped from the center of each reconstructed quadrant, to create the corresponding reconstructed \textit{center}.
Next, the reconstructed \textit{top left, top right, bottom left, bottom right}, and \textit{center} tiles are passed to the  discriminator.
This allows the discriminator to not only ensure that the generator is reconstructing a valid cross-domain result, but also that the central region of a generator output is spatially consistent with its neighborhood data. 
Finally, each input, reconstructed quadrant (multiplied by a desired weight $w_1$), and \textit{center} (multiplied by a desired weight $w_2$) tiles are used to calculate the loss functions for training the network.
Since \textit{center} is the primary tile for reconstruction, we suggest $w_2 > w_1$.
The spatial-consistency results are shown in Fig.~\ref{fig:comparison_spatial_result}(c).

\subsubsection{Implementation Details}
\label{sec:GAN_implementation}

To design neuReGANerator, we have used cycleGAN's architecture implemented by Jun-Yan et al.~\cite{zhu2017unpaired} using PyTorch, as the base structure.
As neurGANerator solves a volume-to-volume translation task, we redesigned cycleGAN to use 3D convolutions in its generator and discriminator. Details of the neuReGANerator training weights and parameters used specifically for our task are provided in the Results section~(Sec. \ref{sec:results}).
To improve training, we replace batch normalization in the discriminator with spectral normalization proposed by Miyato et al.~\cite{miyato2018spectral}, to mitigate mode collapse and exploding gradients.

Due to size limitations of the microscope chamber and  increased scattering of light in thicker biological slices, physical sections of brain samples are very thin (20 - 50 $\mu$m on average).
Therefore, to accommodate the input volume's shallow $z$-depth, we reduce the kernel size of the first and last convolutions in the generator from 7 to 5 and set the stride in the $z$ dimension to be 1 in the discriminator.

We also address an issue with normalizing intensity values of the input volume for neuReGANerator. 
In practice, intensities are normalized between 0 and 1, or -1 to 1, and pushed into the GPU as 16- or 32-bit floating point numbers.
Fig.~\ref{fig:testing_crops} includes the histograms of two raw input volumes of a region of the brain imaged using WFM. 
We can observe that 95\% of the voxels have intensity values less than the mean and a single standard deviation. 
This makes it difficult for the network to distinguish between background and foreground voxels after applying 3D convolutions.
Normalizing input tiles would result in incoherent intensity values across the stitched output volume, as the intensity values from each output tile will be scaled to its local minimum and maximum value. 
Therefore, we perform a non-linear scaling before normalizing and tiling the input volume.
From our experiments, we concluded that scaling the voxels having 0 - 95\textsuperscript{th} percentile intensity values to 0\% - 75\% of the maximum intensity of the volume, and the 96 - 100\textsuperscript{th} percentile intensity values to 76\% - 100\%  of the maximum intensity, yield favourable results. 

As there are multiple solutions to reconstruct fragmented projections and one unique solution to reconstruct a young domain neurite, we have incorporated XDCycleGAN by implementing the extended cycle-consistency loss in the young to old direction to deal with the lossy domain translation.
The extended loss is defined as: 
\begin{equation}
    \mathcal{L}_{xCyc}(G,F,Y) = \mathbb{E}_{y\sim y_{data}(Y)} ~ \delta\Big(G(y),G^\prime(y)\Big) \cdot \| G^\prime(y) - G(y)\|
\end{equation}
where $G^\prime(y) = G(F(G(y)))$. This enforces the following mapping: $\hat{o} \to F(\hat{o}) \to G(F(\hat{o})) \approx \hat{o}$, where $\hat{o} = G(y)$.
This is illustrated in the neuReGANerator training box in Fig.~\ref{fig:pipeline}(a).

\subsubsection{Full Objective}

\label{sec:fullObjective}
Conclusively, the full objective for training the neuReGANerator, with cycle weights  $\Lambda_O$ and $\Lambda_Y$, is:

\begingroup
\setlength\abovedisplayskip{0.15em}
\begin{equation}
\vspace*{-0.25em}
    \begin{split}
        \mathcal{L}_{tile}(G, F, D_O, D_Y) &= \mathcal{L}_{GAN}(F, D_O, y_{tile}, o_{tile})\\
        &+ \mathcal{L}_{GAN}(G, D_Y, o_{tile}, y_{tile})\\
        &+ \Lambda_{O} \mathcal{L}_{cyc}(G,F, o_{tile}) + \mathcal{L}_{hallucination}(o_{tile}, y_{tile})\\
        &+ \Lambda_{Y} \mathcal{L}_{xCyc}(G,F, y_{tile})
    \end{split}
\end{equation}
\setlength\abovedisplayskip{0.15em}
\begin{equation}
\vspace*{-0.25em}
    \begin{split}
        \mathcal{L}_{total} &= w_1\mathcal{L}_{topLeft}(G, F, D_O, D_Y) + w_1\mathcal{L}_{topRight}(G, F, D_O, D_Y)\\
        &+ w_1\mathcal{L}_{bottomLeft}(G, F, D_O, D_Y) + w_1\mathcal{L}_{bottomRight}(G, F, D_O, D_Y)\\
        &+ w_2\mathcal{L}_{center}(G, F, D_O, D_Y)
    \end{split}
\end{equation}
\endgroup
In training neuReGANerator, the network aims to solve:
\begin{equation}
    G^*,F^* = arg\: \min_{G,F}\: \max_{D_O,D_Y}\mathcal{L}(G, F, D_O, D_Y).
    \vspace*{-0.45em}
\end{equation}

\subsection{Visualization}
  \begin{figure*}[th!]
    \centering
    \includegraphics[width = \linewidth]{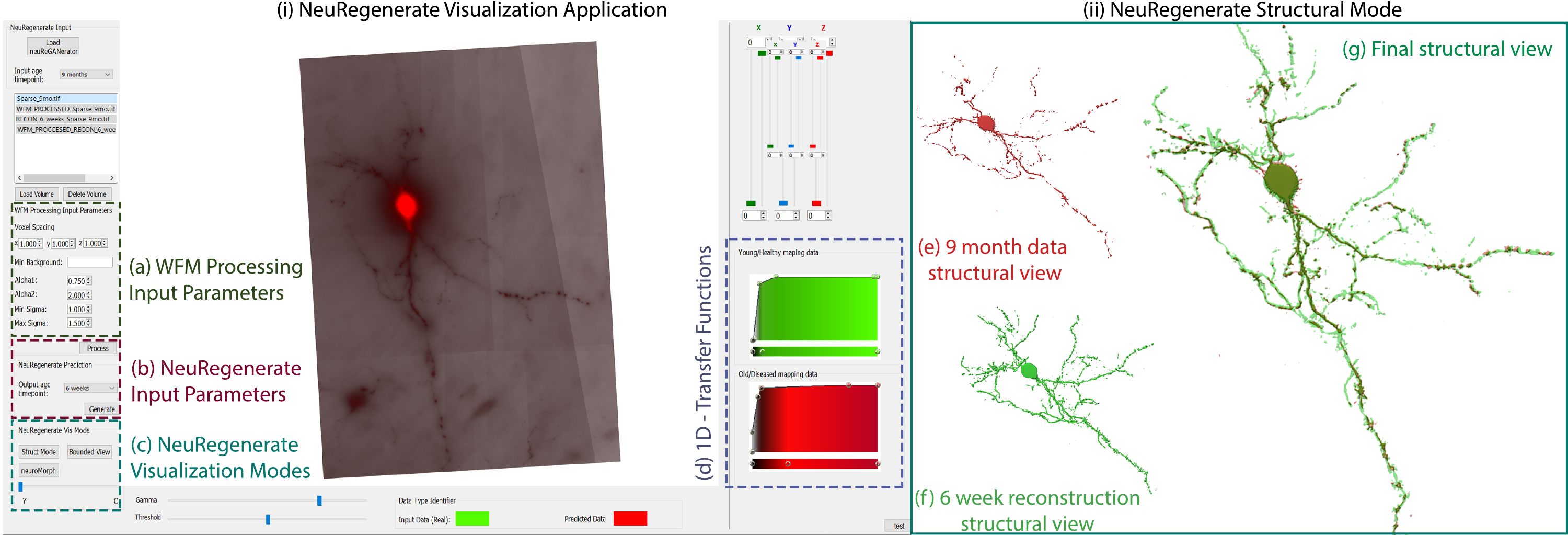}
    \caption{(i) Screenshot of our NeuRegenerate visualization application, and (ii) example of our structural visualization. On loading WFM volume, the user first defines input parameters (a). Next, the input and output age timepoints for the trained neuReGANerator are specified (b). User selects one of two visualization modes (c). User can visualize WF volumes using direct volume rendering and 1D TFs for each age domain (d). 
    The structural visualization of the volume loaded in (i) is shown in (e) and its predicted 6-week neurites is shown in (f). This mode extracts neuronal structures from each domain and renders them in a single view, as shown in (g). }
    \label{fig:visApp}
\end{figure*}
\label{sec:vislualization}
To qualitatively analyze changes in fiber morphology between age timepoints, neuroscientists compare microscopy data using a population of specimens.
Typically, this involves using rudimentary methods such as applying maximum intensity projection to flatten the volume, thresholding intensity values to remove background and noise from the data, and using visual cues to make general observations within the specimen collection.
NeuRegenerate offers a complete visualization framework for studying  changes in micrometer-scale fiber morphology, within a single subject, for two age timepoints-of-interest.
In addition to volume rendering (Sec. \ref{sec:volumeRending}), we design visualization components based on  requirements outlined by neuroscientists in goal \textbf{G2}, for the qualitative analysis of cholinergic fibers in 3D.
Using the predicted structures reconstructed by neuReGANerator, we provide users with two modes, neuroCompare (Sec.~\ref{sec:neurocompare}), and neuroMorph (Sec.~\ref{sec:neuromorph}). 
The underlying assumption in developing these visualizations is derived from domain knowledge (by population analysis) that neuronal fibers undergo fragmentation and loss during aging.
Specifically, neuronal structures of a young brain are more connected and healthier compared to an old brain.
In designing the visualization modes, we also address the challenges of WFM data mentioned in goal \textbf{G3} (Sec.~\ref{sec:Wfm_processing}). 
Finally, to reduce the risk of over-interpretation, we add a label to the application interface that identifies the data (real or predicted) that the user is visualizing. 
A screenshot of NeuRegenerate visualization application is shown in Fig.~\ref{fig:visApp}(i), and a video demonstrating the NeuRegenerate visualization modes is in the \textit{Supplementary Materials}.

\subsubsection{Volume Rendering}
\label{sec:volumeRending}
To render raw WFM data, we use VTK~\cite{schroeder2000visualizing} GPU volume mapper with 1D opacity and color TFs.
By observing the neuroscientists' manipulation of TFs to visualize neuronal structures in WFM volumes, we designed a preset for opacity and color mapping for our application.
We noticed that the preferred mapping is a combination of two linear functions.
For opacity, the mapping consists of: 0 to 0.75 for 0 to 95\textsuperscript{th} percentile of voxel intensity values, and 0.76 to 1.0 for 96 to 100\textsuperscript{th} percentile of the intensity value (on a normalized scale where 1 is the highest opacity value).
Notice that this is the same mapping we use to normalize the input volume to the neuReGANerator in Sec.~\ref{sec:GAN_implementation}.
Likewise, we use the same combination for mapping color values.
Typically, among neuroscientists red and green are most common colors used for visualizing microscopy data. 
Therefore, we chose \textit{black} to \textit{red} to render the old-age domain, and \textit{black} to \textit{green} to render the young-age domain.
The TFs are editable and users can add points on the TFs to create more segments, move points to modify the mapping, and change the color mappings~(shown in Fig.~\ref{fig:visApp}(d)).
For this paper, we use the above mentioned presets to demonstrate our results.
We also include gamma correction and intensity thresholding as additional parameters, as domain experts often use them in their workflow.

\subsubsection{Processing WFM Volumes}
\label{sec:Wfm_processing}
Domain scientists are primarily interested in visualizing neuronal structures. 
However, due to the inherent out-of-focus blurring nature of WFM data, adjusting visualization parameters or directly applying surface rendering techniques to the raw volume is a tedious and ineffective task. 
Current algorithms for tracing and segmenting neuronal structures are best suited for confocal and light-sheet microscopy and they perform poorly on WF data, especially for the genetic labelling and the neurobiological system studied in our task.
Our previous work~\cite{boorboor2018visualization} discusses the challenges of WFM and presents a preprocessing technique for the meaningful rendering of neuronal data.
For NeuRegenerate, we extract vessel-like neurites from WF volumes by first applying our gradient-based distance transform function~\cite{boorboor2018visualization} to suppress background voxels.
Next, given its ability to  exhibit a high and uniform response for vessels with variable morphology, we apply Jerman's enhancement filter~\cite{jerman2016enhancement} to the processed volume.
Parameters for the enhancement filter (voxel spacing and sigma) are specified by the user on loading the input volume.
Voxel spacing is determined from the microscope meta-data and sigma is the range of the neurite cross-sectional diameter, entered based on the user's domain knowledge of the data.
The outputs from the vesselness filter are the best responses to the enhancement function using a multiscale Hessian matrix.
However, neurites with weak intensity values exhibit a low response in regions of strong blurring or presence of high-intensity neurites, and often get thresholded in the process of removing noise from the filter response. 
To resolve this, we adopt a multiscale intensity approach for recovering Hessian gradients of weak neurites.
Intensity scales are determined using the piece-wise opacity TF adjusted by the user.
Finally, to remove artifacts from the filter response, users interactively specify a minimum threshold value using a slider in the application interface.

\subsubsection{neuroCompare Mode}
\label{sec:neurocompare}

In this mode, we simultaneously visualize neuronal structures from two age domains using structural and bounded views.

\noindent
\paragraph*{\textbf{Structural View}} We render neuronal fibers from the young-age volume as hollow green translucent structures encapsulating the fragmented projections from the old-age volume, rendered in red.
The iso-surface of the neuronal structures are extracted from the processed WFMvolume using VTK marching cubes filter. 
Fig.~\ref{fig:visApp}(ii) shows an example of our structural representation.
The red structure in Fig.~\ref{fig:visApp}(e) is the structural rendering of a 9-month neuronal data (processed from the WF volume shown in Fig.~\ref{fig:visApp}(i)), and the green structure in Fig.~\ref{fig:visApp}(f) is the structural rendering of its corresponding predicted 6-week structure.
Fig.~\ref{fig:visApp}(g) demonstrates the final combined structural representation.
Using this mode, users can observe the morphological profile of the original 9-month neuronal data (in red), in comparison to its possible younger state (in green).

\begin{figure}[t]
    \centering
    \includegraphics[width = \linewidth]{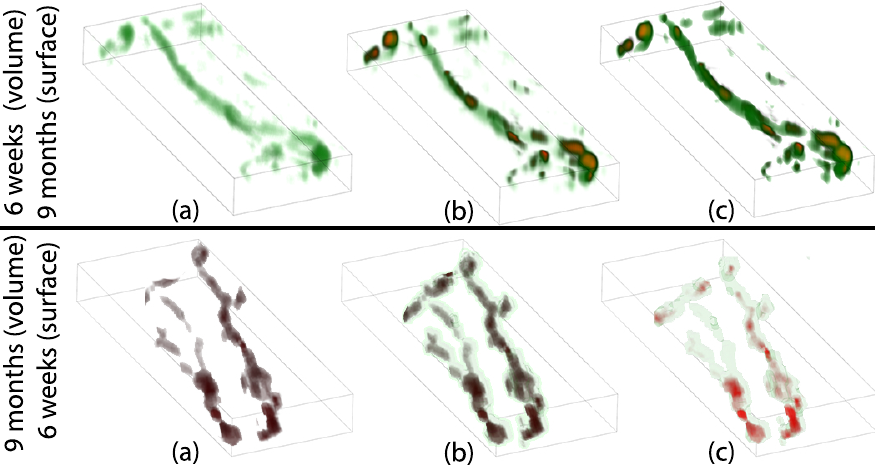}
    \caption{Example of our bounded view visualization. For each direction, a volume rendering of the bounded raw input WF data is shown in (a), followed by the surface rendering of its predicted structure in (b). Parameters such as intensity thersholding and gamma correction can be adjusted for the volume, as demonstrated in (c).  }
    \label{fig:boundView}
\end{figure}

\noindent
\paragraph*{\textbf{Bounded View}} 
This view is designed to allow exploration of the input volume with respect to the predicted neuronal data.
The input volume is rendered using direct volume rendering with TFs and enhancement parameters, whereas the predicted neuronal data is rendered as iso-surfaces.
If the input volume belongs to the young domain, bounds from the thresholded vesselness filter is used to mask the foreground from the background voxels. Conversely, vesselness bounds of the predicted young structures are used to mask voxels if the input volume belongs to the old domain. 

The utility of such a representation is that volume rendering parameters, such as thresholding and gamma correction, are widely used by neuroscientists to observe features of neuronal data in the microscopy volume.
Since the volume reconstructed by neuReGANerator is solely a prediction, allowing its visual exploration can carry a risk of over-interpretation by the user.
Therefore, for this view, we restrict volume visualization of the predicted data to iso-surfaces, and only the real input data to be explored using volume rendering and TFs. 
Fig.~\ref{fig:boundView} shows examples of the bounded representation for both age domains and their predicted data.

\subsubsection{neuroMorph Mode}
\label{sec:neuromorph}

\begin{algorithm}[t]
\SetAlgoLined
\DontPrintSemicolon 
\KwIn{Static voxels s($O$), and dynamic voxels d($Y$) }

$I(Y) \gets  d(Y) \cap s(O)$ \;

  H $\gets minHeap()$\;
  \lForEach{$\alpha \in d(Y)$}
  {$\alpha.dist \gets \infty$, $\alpha.status = notAssociated$ }
  \ForEach{$\alpha \in I(Y)$}{
    \ForEach{neighbour $\beta$ of $\alpha$, where $\beta \in d(Y)$}{
      $\beta.dist = eucledianDistance(\alpha_{x,y,z},\beta_{x,y,z} )$\;
      $\beta.status = associated$ \;
      $\beta.path \gets \alpha$ \;
      $H.push(\beta)$ \;
      }
     }
    \While{$H$ \textbf{is not empty}}
    {
        $\alpha = H.pop()$ \;
        $\alpha.status = associated$ \;
        \ForEach{neighbour $\beta$ of $\alpha$, where $\beta \in d(Y)$ and $\beta.status \neq associated $}{
            \textit{d} $ = \alpha.dist + eucledianDistance(\alpha_{x,y,z},\beta_{x,y,z} )$\;
            \uIf{$\beta.status = notAssociated$}
            {
                $\beta.dist = d$, $\beta.status = queued$\;
                $\beta.path \gets \alpha.path + \alpha$ \;
                $H.push(\beta)$ \;
            }
            \uElseIf{$\beta.status = queued$}{
                \uIf{\textit{d} $ < \beta.dist $}{
                    $\beta.dist = d$\;
                    $\beta.path \gets \alpha.path + \alpha$ \;
                }
            }
            
        }
  }
  
 \caption{Path between dynamic voxels to static voxels}
 \label{alg:morphEq}
\end{algorithm}

\begin{figure}[t]
    \centering
    \includegraphics[width = \linewidth]{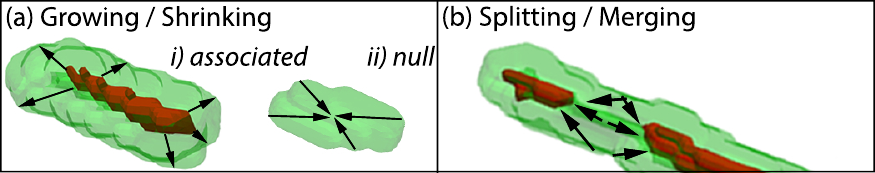}
    \caption{The path for dynamic voxels follow either (a) growing/shrinking or (b) splitting/merging action.}
    \label{fig:neuroTopologies}
\end{figure}

The percentage fragmentation of fibers vary for different sub-regions of the brain. 
Using neuReGANerator, neuroscientists are interested to study the pattern of neurite fragmentation and deformation, across regions of the brain, during the neurodegradation process.
Though this can be observed in the structural mode, the high density of neuronal data in the volume can make this analysis task overwhelming.
To this end, for a more illustrative and interactive understanding of the morphological changes, outlined in goal \textbf{G2}, we develop a morphing technique called neuroMorph.
NeuroMorph is a progressive algorithm that computes meaningful transformations between a source and a target volume, specifically taking into account the tubular-shaped morphology (vesselness) of the neuronal fibers.
Following the hypothesis that neurites undergo fragmentation and thinning during aging, we classify voxels of a neurite as \textit{static} - neurite voxels that are in both the young and old domain; or \textit{dynamic} - neurite voxels that are only in the young domain.
For each neurite, we assign a path from its dynamic voxels to the boundary of its correpoding static voxels.
We then perform a linear transformation along the assigned paths, to reconstruct the neurites' intermediate structures.
It is important to note here that the in-between volumes generated using neuroMorph serve only as a visualization for the temporal deformation of the neurites across the specified age-timepoints.
For an accurate representation of the in-between reconstructions, users will have to train the neuReGANerator for the additional age-timepoints.
To avoid over-interpretation, the interface displays a prediction data warning label while the user scrolls through the reconstructed structures.

The path for each \textit{dynamic} voxel is based on one of the following topological events: growing/shrinking or splitting/merging~(Fig.~\ref{fig:neuroTopologies}).
A special \textit{null} case exists in the growing/shrinking event in which a young neurite does not have a corresponding old neurite~(Fig.\ref{fig:neuroTopologies}(ii)).
Algorithm~\ref{alg:morphEq} describes our path assignment approach. 
Since neurites can be processed in parallel, we first extract the bounds of each neurite from the structural representation of the young and old volumes and pair intersecting bounds for computing the morphing trajectory.
For each paired neurite, we determine the set of voxels from the young domain that intersect with the iso-surface of its corresponding old domain as static. 
The remaining voxels in the young domain are labelled as dynamic.
Next, we use fast marching~\cite{sethian1996fast} to construct a path from the set of static voxels to each dynamic voxel.
A path exchange is performed if a dynamic voxel encounters a shorter path to a static voxel.
For \textit{null} case neurites, we determine its mass center voxel as its static voxel.

\begin{figure*}[t!]
\includegraphics[width=\linewidth]{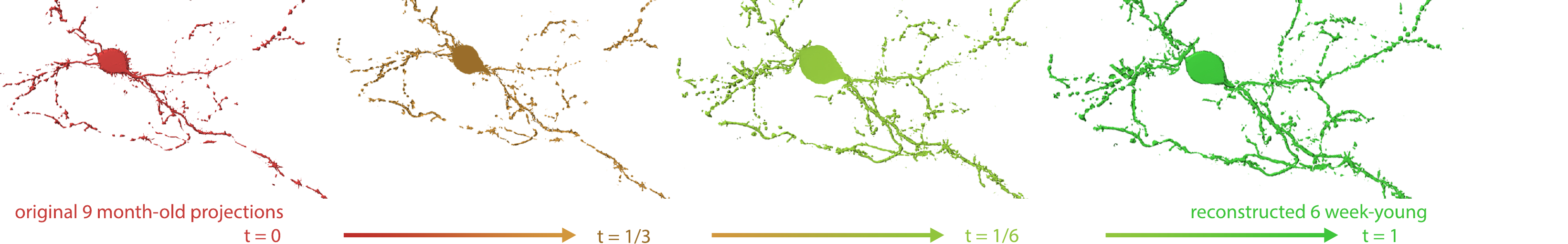}
\vspace{-6mm}
  \caption{NeuRegenerate aids neuroscientists in visualizing structural changes that occur within a specimen brain, across age: for a diseased mouse data~(left-most structure), we are able to predict and reconstruct its healthy neuronal fibers at a younger age~(right-most structure). The two in-between structures are generated using neuroMorph, that allows users to interactively visualize the neurodegeneration process.}
	\label{fig:teaser}
 \end{figure*}

To reconstruct intermediate volumes, we perform a linear transformation with parameter $\sigma \in [0,1]$, over a regular grid, along the assigned path of the dynamic voxels, starting from the input age-domain to the output domain.
For each dynamic voxel of a neurite, the resultant intermediate volume in the $Y \to O$ direction is the $1 - ceil(\sigma \times pathLength)$ voxels along its assigned path, and $floor(\sigma \times pathLength)$ voxels for the $O \to Y$ direction.
Fig.~\ref{fig:teaser} demonstrates the result of our neuroMorph algorithm for the neuronal data shown in Fig.~\ref{fig:visApp}, with intermediate reconstructions from its input \textit{old} neuronal structures to its  predicted \textit{young} structure, for $\sigma = [0, \frac{1}{3}, \frac{1}{6}, 1]$.

\begin{figure*}[t!]
  \includegraphics[width=\textwidth,keepaspectratio]{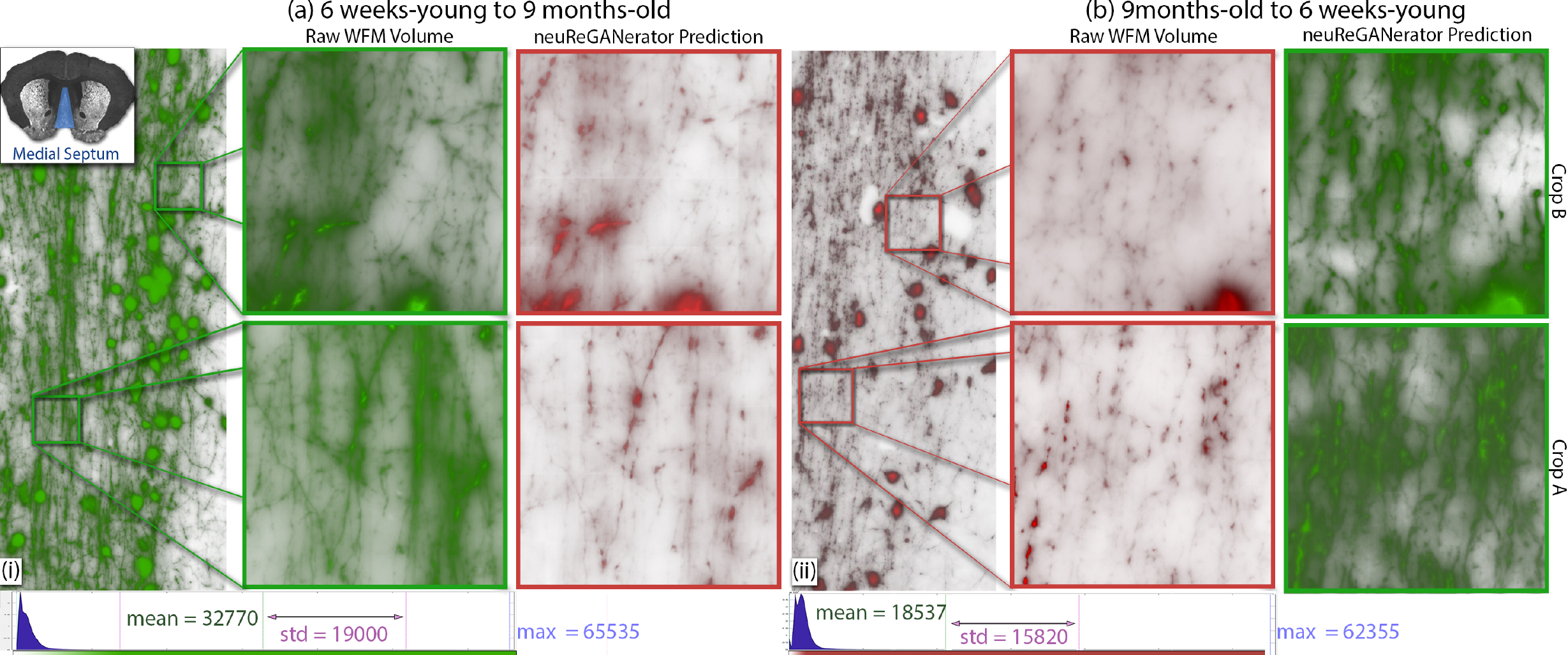}
  \vspace{-4mm}
  \caption{The results of neuReGANerator using 6-week and 9-month cholinergic neurons in the medial septum of a mouse brain. We focus on two regions (Crop A and B)  of dimensions 500$\times$500$\times$20. The green volume in (i) is a testing 6-week input.  The red volumes in (a) are the corresponding 9-month predicted reconstructions of the cropped regions. The red volume in (ii) is a testing 9 month-old input. Similarly, the green volumes in (b) are the corresponding 6 week-old predicted reconstructions. The plots at the bottom are histograms of the respective input volumes. The histograms demonstrate the skewness of the voxel intensity distribution, thus making visualization of WFM volumes a challenge.
  }
  \label{fig:testing_crops}
\end{figure*}

\begin{figure*}[t]

  \includegraphics[width=\linewidth,keepaspectratio]{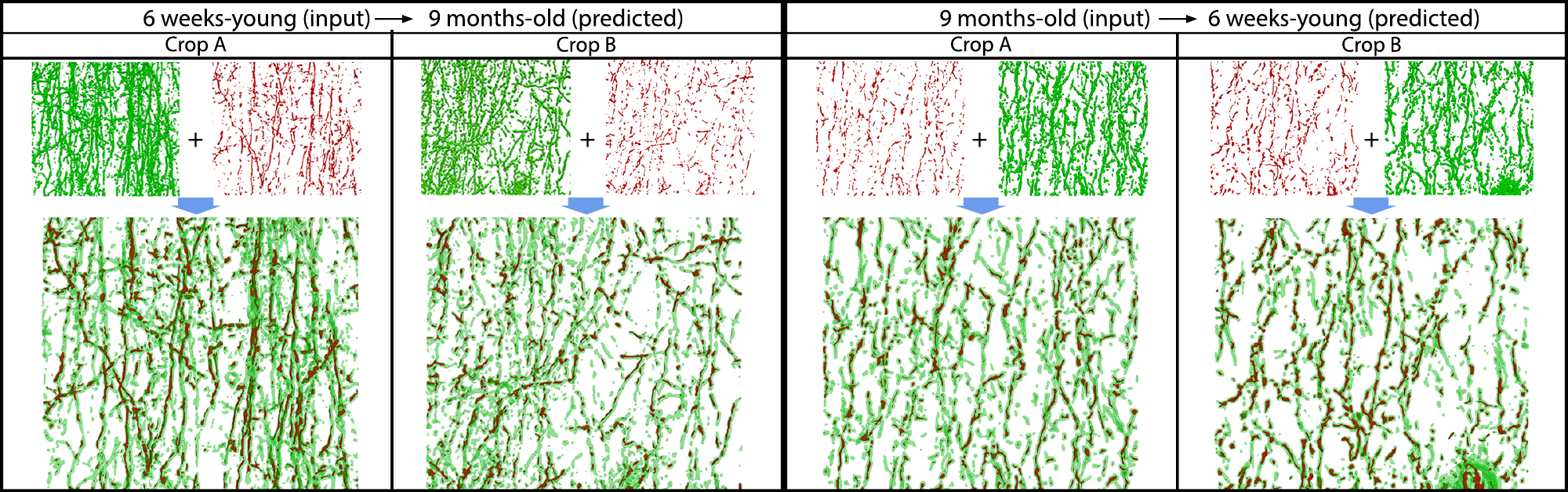}
  \vspace{-6.5mm}
  \caption{The structural view visualizations for the cropped regions (Crop A and Crop B) shown in Fig.~\ref{fig:testing_crops}  }
  \label{fig:structural_results}
\end{figure*}

\begin{figure*}
  \includegraphics[width=\linewidth,keepaspectratio]{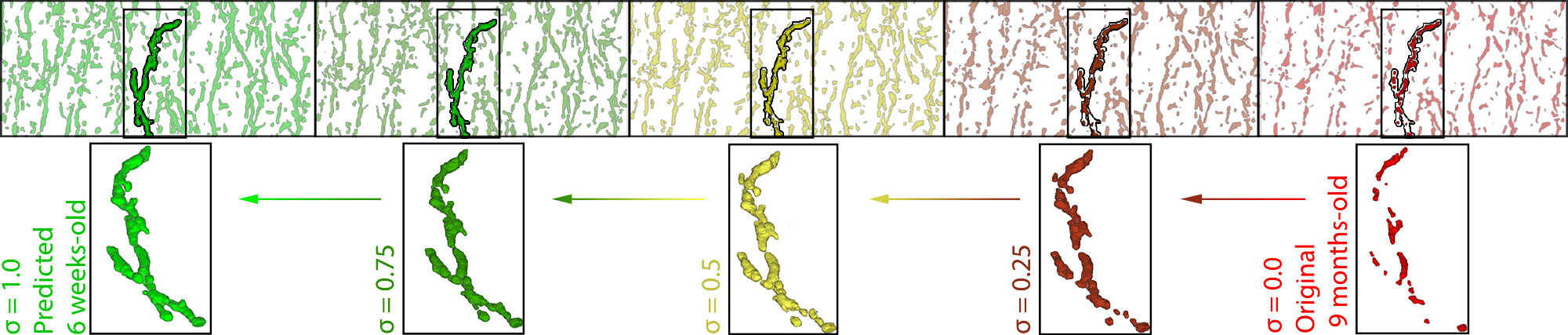}
  \vspace{-6.5mm}
  \caption{Our neuroMorph visualization for the 9 months to 6 weeks Crop B example.}
  \label{fig:morph_results}
\end{figure*}

\section{Results and Evaluations}
\label{sec:results}

We demonstrate NeuRegenerate using 6 week~(6wk) and 9 month~(9mo) timepoints of cholinergic fibers of mice specimens.

\paragraph*{\textbf{Biological Prep}} 
Samples were generated from a transgenic mouse line with a tau-enhanced green fluorescent protein under the control of the ChAT promotor (ChAT-tau-eGFP) labeling cholinergic neurons and extensions.
At the appropriate age mice brain tissue were serially sectioned at 20$\mu$m thickness and imaged using an Olympus~VS-120 WF microscope with a numerical aperture of 0.95 at $40\times$ magnification, and $xy$ resolution of 0.325$\mu$m/pixel and $z$ spacing of 1~$\mu$m/pixel.

\paragraph*{\textbf{NeuReGANerator Input}}
12 mice specimens were used for our experiments, six from each age domain (6wk and 9mo).
In total, we trained NeuReGANerator using 6724 tiles per domain.
We determined that tiles of size 128 $\times$ 128 with $\Delta =$ 32 best captured the neuronal information for spatial-consistency.
Thus, input data for training and testing were cropped into tiles of dimensions 
 $ 192 \times 192 \times 20$.
The testing tiles had an overlap of 64 voxels in the $x$-$y$ directions. 
For all results shown in this paper, NeuReGANerator was trained for 20 epochs (134480 iterations) with a learning rate of 0.0002 for the first 10 epochs, followed by a step decay with a multiplicative factor of 0.1 for the last 10 epochs, for both the generator and discriminator.
We used $\lambda_O = 10$ and $\lambda_Y = 10$ for the hallucination loss,  $\Lambda_O = 10$ and $\Lambda_Y = 10$ for $\mathcal{L}_{cyc}$, and ($w_1 = 0.25$, $w_2 = 0.5$) for the spatial-consistency weights.

\par
Fig.~\ref{fig:testing_crops} shows the output volumes predicted by NeuReGANerator using input volumes from the medial septum region, for each age domain.
As described in Sec.~\ref{sec:vislualization}, visualizations in green represent the young (6wk) domain and red represents the old (9mo) domain. 
Since the medial septum is very large and dense, we extract two crops from each domain, \textit{Crop~A} and \textit{Crop~B}, to present detailed results of our visualizations.
For each crop, Fig.~\ref{fig:structural_results} shows their respective structural view visualization using neuroCompare and Fig.~\ref{fig:morph_results} demonstrates the intermediate volumes reconstructed using neuroMorph.

We presented NeuRegenerate to neuroscientists, who are experts in the study of cholinergic systems across lifespan, to evaluate our framework.
Sec.~\ref{sec:domain_eval} provides an account of their evaluation, followed by a case study in Sec.~\ref{sec:case_study}, that resulted in a novel hypothesis towards an early-onset structural change in aging. A discussion on using NeuRegenerate by the experts is presented in Sec.~\ref{sec:discussion}.  

\subsection{Domain Expert Evaluation}
\label{sec:domain_eval}

We evaluated NeuRegenerate using a region of the brain we have studied extensively using population data.
The input samples (shown in Figs.~\ref{fig:testing_crops}, \ref{fig:structural_results}, and \ref{fig:morph_results}) are prepared using a technique that provides a wide-spread labeling of most fibers of our cell type of interest, allowing for broad, exploratory analysis.
By closely examining 20 input volumes, we have determined that the predicted volume and visualization plausibly exhibit age-related changes to fiber morphology and health.
However, given that the nature of this work lacks the ability to validate prediction results using groundtruth, we performed a quantitative analysis using fiber density measure~\cite{ballinger2019mecp2}.
This metric is widely used by neuroscientists to measure the density of a neuronal network that manifests the health of communication in the brain and is computed as the ratio of neurons to empty volume in a region.
We computed the density measure on the training  datasets (at 6wk and 9mo), as well as for the testing data set and their reconstructed counterparts.
Comparing the percentage difference of fiber density across groups reveals that despite individual variability that is expected across samples (variability in absolute fiber density), neuReGANerator is able to perform well on an individual basis (see Table~\ref{tab:quantResults}).

\begin{table}[h]
\centering
\caption{The fiber density measure for 6-week and 9-month show that neuReGANerator preserves overall fiber density difference. }
\label{tab:quantResults}
\begin{tabular}{|l|l|l|l|} 
\cline{2-4}
\multicolumn{1}{l|}{}                                                   & \multicolumn{2}{l|}{~ ~ Fiber Density} & \multirow{2}{*}{\begin{tabular}[c]{@{}l@{}}Percentage\\~Difference\end{tabular}}  \\ 
\cline{2-3}
\multicolumn{1}{l|}{}                                                   & ~ 6wk   & ~ 9mo                      &                                                                                   \\ 
\hline
\begin{tabular}[c]{@{}l@{}}Training Dataset \\6wk and 9mo\end{tabular}  & 42.5\% & 6.8\%                     & ~ 82.7\%                                                                           \\ 
\hline
\begin{tabular}[c]{@{}l@{}}6wk (input)\\to 9mo (predicted)\end{tabular} & 27.1\% & 3.7\%                     & ~ 85.5\%                                                                           \\ 
\hline
\begin{tabular}[c]{@{}l@{}}9mo (input)\\to 6wk (predicted)\end{tabular} & 39.8\% & 5.8\%                     & ~ 86.5\%                                                                           \\
\hline
\end{tabular}
\end{table}

One feature that we observed using neuroCompare was the extent of fiber thinning that occurs in aging.
Coupled with neuroMorph, this provided a unique comparative insight into the proportion of thinning and fragmentation that contribute towards the overall fiber loss, quantitatively determined using fiber density measures.
This is novel for our field and this type of within subject visualization is beneficial in conveying our findings to expert and non-expert audience.

A significant utility of NeuRegenerate is to evaluate predicted changes to fiber morphology in small, segregated populations of neurons that are functionally important to behavioral sequences.
For instance, spatial memory deficits are a known feature of aging across species. 
To this end, we conducted an experiment using an isolated 9mo sample as shown in Figs.~\ref{fig:visApp} and \ref{fig:teaser}.
Since input volume contained only the projections of an isolated neuron, we were able to verify the connections of the fragmented fibers that neuReGANerator reconstructed, as in a 6wk state.
Using neuroCompare and neuroMorph to visualize the change in morphology between the predicted 6wk structure and the real 9mo sample, we were able to observe degradation patterns relative to attributes such as fiber thickness, distance from the cell-body, branching morphology, and brain region – something that we have not yet been able to do given methodological limitations. 
This is specifically important for understanding how cells that are functionally important for performing specific spatial memory tasks change as a function of age will be crucial to our understanding of age-related cognitive impairment.

\subsection{Case Study: Degeneration in the Cortex}
\label{sec:case_study}
Following the analysis of the medial septum, we conducted a case study on the cortex -- a region of the brain where the cholinergic fibers are more resilient in aging and change to fiber morphology across 6wk and 9mo samples is not expected.
We used NeuRegenerate and trained neuReGANerator for the cortex of 6wk and 9mo mouse specimens, using parameters and preparation method mentioned above, and visualized the resulting change in fiber morphology for individual specimens.
On observing the raw volumes, similar to our knowledge and expectation of cortical fiber morphology, we found that the predicted results showed no fragmentation between the real input and predicted volumes.
Moreover, the fiber density of this region resulted in around 5\% loss of fiber density which falls within the expected margin of biological variability, thus reassuring structural resilience.

\begin{figure}
\centering
  \includegraphics[width=\linewidth,keepaspectratio]{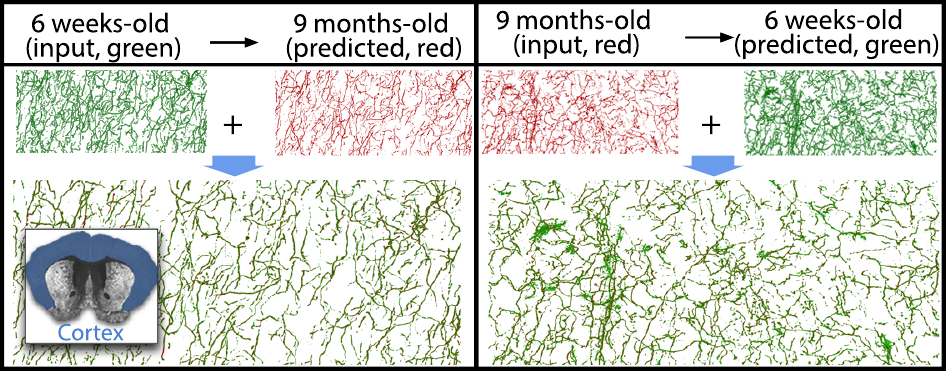}
  \vspace{-6mm}
  \caption{The structural view of the cropped cortical region. 
  }
  \label{fig:cortex_results}
  \vspace*{-1.5em}
\end{figure}

However, as we proceeded to evaluate using neuroCompare, we were presented with a surprising result: the cortical fibers were thinner in 9mo samples than in the 6wk samples (shown in Fig.~\ref{fig:cortex_results}). 
This subtle difference is almost impossible to observe with just population analysis and is not significantly captured using the primary fiber density analysis. 
Upon closer observation of population data images, we were able to corroborate independently by measuring the average thickness of the fibers in the collected samples.

This case exemplifies the utility of NeuRegenerate for biological research as it allowed for a novel observation towards an early, detectable feature of morphology change in lifespan. 
While it is true that the predicted outputs of this framework are not accurate biological structures, they allow for the generation of new questions and hypotheses and fill an obvious gap in methodology that can only be overcome with significant advances in long term \textit{in-vivo} deep tissue imaging techniques.

\subsection{Discussion}
\label{sec:discussion}
NeuRegenerate fills critical gaps in methodology and analysis in the neuroscience field and, for the first time, has allowed a within subject prediction of regional fiber changes with age. 
Any previous attempts toward this endeavor have required the use of population data which is not amenable to direct overlap and subsequent evaluation of images in this way.
It has provided an opportunity for effective brain-wide exploratory analyses. 
Typically, because of the labor-intensive nature of our current image analysis workflow, we study regions where we hypothesize changes in fiber morphology because of obvious changes in functional output of the region (example, behavior). 
The limitation with such an approach is that we do not know what we are missing and have not yet been able to adequately explore all regions of the brain. 
It is well established that there exists broad regional and cell type specific heterogeneity in degradation across lifespan. 
NeuRegenerate is the first framework of its kind to break down the within-subject barrier and allow for generation of new hypotheses. 
By studying a brain-wide predicted fiber morphology change as output, we can begin to predict a progression of cell-type X region specific vulnerability across lifespan. 
Though this paper focuses on lifespan, this framework will be instrumental in studying the relationship between brain structure and function in normal aging and using it to compare to other critical areas of research such as stress, drug addiction, trauma, and other disease states.

\section{Limitations}
\label{sec:limitations}

NeuRegenerate is designed based on the current key neuroscience postulation that fragmentation and reduction in fiber density (vessel thickness) are significant features of change in fiber morphology.
Specifically, to avoid meaningless structures and artifacts in the prediction outcome, the hallucination loss inhibits  how  much  change  is allowed  when transitioning from young to old or vice-versa.
Likewise, the intermediate volume reconstruction method in neuroMorph caters only to growing/shrinking and splitting/merging of vessel-like neurites.
This limits NeuRegenerate scope to visualize and hypothesize supplemental fiber morphology variations beyond thickness, decay, and connective growth (for instance, branching) that could possibly occur in other neurological systems.
Moreover, the conservative nature of the hallucination loss poses an interesting dilemma: is the network hindered from reconstructing previously unobserved changes that can be discovered using NeuRegenerate? 
Notwithstanding that relaxing this loss may make it difficult to distinguish true from hallucinated features.
Thus, expanding this work scope for additional morphological changes will require designing a more novel hallucination loss and a more extensive intermediate volume reconstruction algorithm for neuroMorph. 

The lack of real biological groundtruth correspondence was a challenge in designing our framework. 
While NeuRegenerate provides great visualizations of neuronal structures, it is unintended for diagnosis or ground truth. 
All findings using NeuRegenerate requires verification by domain experts using either evaluation metrics that can estimate the plausibility or through additional medical experiments. 
Coupled with the visualization tools, this framework  can greatly aid researchers in developing hypotheses.
If ground truth is made available, NeuRegenerate can be extended to become better suited for diagnosis, however, currently stage ground truth is unavailable.

\section{Conclusion and Future Work}
We have presented NeuRegenerate, a novel framework that predicts and visualizes the health and density of neuronal fibers, across discrete age timepoints, for an individual specimen. 
Advances in connectomics research have allowed neuroscientists to gain insights into neuron morphology and connectivity and have been instrumental in furthering the understanding of human brain diseases.
Our work helps neuroscientists visualize prediction of fiber health (future  or past) by  providing a necessary within-subject estimation of morphological changes in projections. 
NeuRegenerate consists of a deep-learning component (neuReGANerator) that predicts the translation of fiber structure across trained discrete age-timepoints, and two visualization modes (neuroCompare and neuroMorph) to study the changes between the input and reconstructed volumes. 

In addition to the extensions described in Sec.~\ref{sec:limitations}, we plan to design a single network that can simultaneously learn fiber reconstructions across multiple brain regions, for each age timepoint. Moreover, neuReGANerator focus has been on neurite reconstructions, however, the loss of cell bodies is a known feature in aging and cognitive decline. Thus, we plan to extend neuReGANerator to predict cell-body count change and reconstruct predictions for the age timepoints.

\ifCLASSOPTIONcompsoc
  \section*{Acknowledgments}
\else
  \section*{Acknowledgment}
\fi

This research was supported in part by NSF grants CNS1650499, OAC1919752, ICER1940302, and IIS2107224	and by the Intramural Research Program of the NIH, NINDS and NIMH.

\ifCLASSOPTIONcaptionsoff
  \newpage
\fi



%
\bibliographystyle{IEEEtran}
\bibliography{IEEEabrv,references}


%
\begin{IEEEbiography}[{\includegraphics[height=1.25in,width=1in,keepaspectratio]{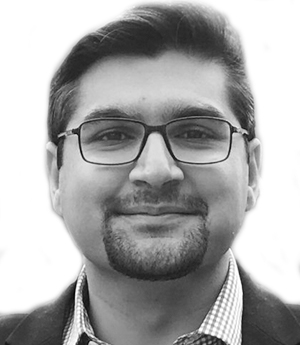}}]{Saeed Boorboor}
is currently persuing a PhD degree in  Computer Science at Stony Brook University. He received his BSc Honors in Computer Science from School
of Science and Engineering, Lahore University of Management Sciences, Pakistan. His research interests include scientific visualization, biomedical imaging, and computer graphics.
\end{IEEEbiography}

\begin{IEEEbiography}[{\includegraphics[keepaspectratio]{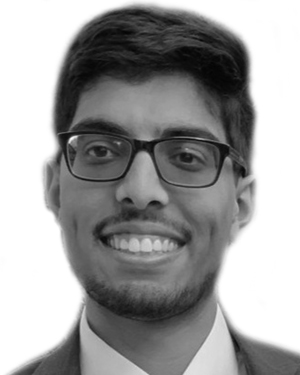}}]{Shawn Mathew}
is a Computer Science PhD Candidate at Stony Brook University. He received his BS in Computer Science from Willaim E. Macaulay Honors College, CUNY City College. His research interests include computer vision, medical imaging, deep learning, and machine learning.
\end{IEEEbiography}

\begin{IEEEbiography}[{\includegraphics[height=1.25in,width=1in,keepaspectratio]{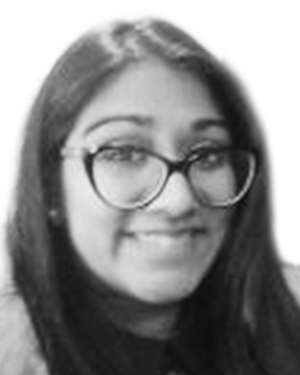}}]{Mala Ananth}
is a Postdoctoral Research Fellow at the National Institute of Neurological Disorders and Stroke. She received her B.S. in Biology from Stony Brook University in 2011 and worked as a Research Assistant at Brookhaven National Laboratory through 2013. Dr.Ananth completed her PhD in Neuroscience from Stony Brook University in 2019. Her research investigates the heterogeneity of cell types in age-related cognitive decline.
\end{IEEEbiography}

\begin{IEEEbiography}[{\includegraphics[height=1.25in,width=1in,keepaspectratio]{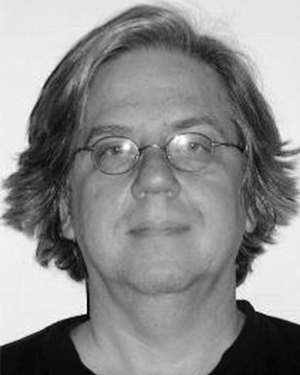}}]{David Talmage}
 is currently a Senior Scientist at the National Institute of Mental Health, NIH. He received a BA in Biology from the University of Virginia, a PhD in Genetics from the University of Minnesota and post-doctoral training at the Rockefeller University and at Harvard Medical School.  Dr. Talmage has authored or co-authored over 70 peer-reviewed articles that have been cited nearly 4000 times.
\end{IEEEbiography}

\begin{IEEEbiography}[{\includegraphics[height=1.25in,width=1in,keepaspectratio]{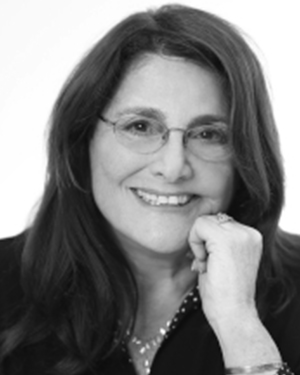}}]{Lorna W. Role}
 is the Scientific Director and Senior Investigator at NINDS, NIH. She obtained an A.B. in applied mathematics and Ph.D. in physiology from Harvard University, and postdoctoral training in pharmacology at Harvard Medical School and Washington University School of Medicine. She became an assistant professor at Columbia University in 1985 and became a professor before moving to Stony Brook University in 2008. There, she served as a SUNY Distinguished Professor and Chair of the Department of Neurobiology and Behavior, and co-director of Neurosciences Institute. She has earned many awards and honors, including Fellow of the American Association for the Advancement of Science (2011) and Fellow in the American College of Neuropsychopharmacology (2009).
\end{IEEEbiography}

\begin{IEEEbiography}[{\includegraphics[keepaspectratio]{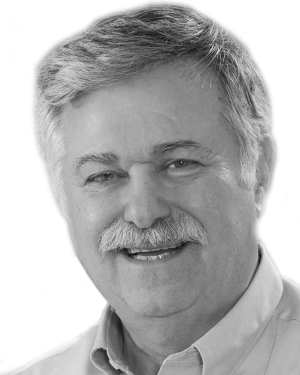}}]{Arie E. Kaufman} is a Distinguished Professor, Director of Center of Visual Computing, and Chief Scientist of Center of Excellence in Wireless and Information Technology at Stony Brook University. He served as Chair of Computer Science Department, 1999-2017. He has conducted research for $>$40 years in visualization and graphics and published $>$350 refereed papers. He was the
founding Editor-in-Chief of IEEE TVCG, 1995-98. He is an IEEE Fellow, ACM Fellow, National Academy of Inventors Fellow, recipient of IEEE Visualization Career Award (2005), and inducted into Long Island Technology Hall of Fame (2013) and IEEE Visualization Academy (2019). He received his PhD in Computer Science from Ben-Gurion University, Israel (1977).
\end{IEEEbiography}




\end{document}